\documentclass{article}

 \usepackage[preprint]{corl_2026} 

\usepackage{microtype}
\usepackage{graphicx}
\usepackage{subcaption}
\usepackage{booktabs} 
\usepackage{xspace}
\usepackage[table]{xcolor}
\usepackage{listings}

\lstdefinelanguage{YAML}{
  keywords={true,false,null},
  keywordstyle=\color{black},
  sensitive=false,
  comment=[l]{\#},
  commentstyle=\color{gray}\ttfamily,
  stringstyle=\color{black}\ttfamily,
  morestring=[b]',
  morestring=[b]"
}

\usepackage{hyperref}

\usepackage{amsmath}
\usepackage{amssymb}
\usepackage{mathtools}
\usepackage{amsthm}

\usepackage[capitalize,noabbrev]{cleveref}

\title{Agent-driven Long-tail Simulation for \\ Autonomous Driving}

\author{
    Junru Gu$\phantom{}^{1}$
    \And
    Lijin Yang$\phantom{}^{2}$
    \And
    Jianing Huang$\phantom{}^{2}$
    \And
    Shu Liu$\phantom{}^{2}$
    \And
    Zhongzhan Huang$\phantom{}^{2}$
    \And
    Hang Zhao$\phantom{}^{1}\phantom{}^{\text{†}}$
    \vspace{15pt}
    \\
    $\phantom{}^1$IIIS, Tsinghua University
    \hspace{12pt}
    $\phantom{}^2$Bosch Research\hspace{5pt}
}

\begin{document}
\maketitle

\newcommand{\dataset}{SemanticPlan\xspace}

\newcommand{\goalmetric}{Goal Compl.\xspace}
\newcommand{\collisionmetric}{No Col.\xspace}
\newcommand{\drivablemetric}{Lane Compa.\xspace}
\newcommand{\longtailmetric}{Long-tail score\xspace}

\newcommand{\scorecollision}{\ensuremath{S_{\text{coll}}}\xspace}
\newcommand{\scoreregion}{\ensuremath{S_{\text{region}}}\xspace}
\newcommand{\scorehonk}{\ensuremath{S_{\text{honk}}}\xspace}
\newcommand{\cmark}{\checkmark}
\newcommand{\xmark}{\ensuremath{\times}}

\begin{abstract}
Evaluating autonomous driving systems in closed-loop settings requires realistic and interactive simulation, yet existing simulators largely rely on log replay or rule-based agents, limiting behavioral diversity and long-tail coverage.
We propose an agent-driven simulation framework in which surrounding road participants are controlled by instruction-following large language models through a structured action interface, enabling intentional and reactive behaviors while preserving physical plausibility.
Furthermore, we introduce SemanticPlan, a benchmark of closed-loop planning in long-tail and semantically rich scenarios that augment real nuPlan scenes with multiple interactive agents following diverse language instructions. Evaluation results show that state-of-the-art planners still struggle to consistently achieve safe and effective task completion, suggesting that these long-tail scenarios remain challenging.
\footnote{Project page: \url{https://tsinghua-mars-lab.github.io/SemanticPlan}}
\begingroup
\renewcommand{\thefootnote}{\textdagger}
\footnotetext{Corresponding to: hangzhao@mail.tsinghua.edu.cn}
\endgroup
\end{abstract}


\section{Introduction}

To accurately and comprehensively evaluate autonomous driving systems, a realistic and interactive simulation environment is essential.
In particular, surrounding road participants should be able to respond dynamically to the actions of the autonomous system, enabling meaningful closed-loop interaction.

Existing simulation frameworks predominantly rely on log replay or simple rule-based models to simulate the behavior of surrounding road participants~\cite{li2024choose_review}.
While these approaches are efficient, they fail to capture the complexity and diversity of real-world human behaviors and cannot adequately respond to the actions of the tested autonomous system.
For example, in closed-loop simulation settings, nuPlan~\cite{caesar2021nuplan,karnchanachari2024nuplan} employs log replay for pedestrian motion and IDM~\cite{treiber2000idm} for surrounding vehicles, resulting in limited interactivity and realism.

In addition, long-tail scenarios are inherently rare in real-world driving datasets. Although simulators such as CARLA~\cite{dosovitskiy2017carla} and interPlan~\cite{hallgarten2024interplan} attempt to address this issue by introducing synthetic data or additional road participants, they still rely on rule-based motion logic. As a result, they do not reflect realistic human behaviors and cannot adequately cover diverse and complex motion trajectories.

To address these issues, we propose an agent-driven simulation framework, where specific road participants are controlled by instruction-following agents instead of predefined rules.
This approach enables the simulation of complex and rare scenarios, where each road participant behaves as a realistic human with individual intentions and reactions.

\begin{figure}[t]
    \centering
    \includegraphics[width=0.7\linewidth,]{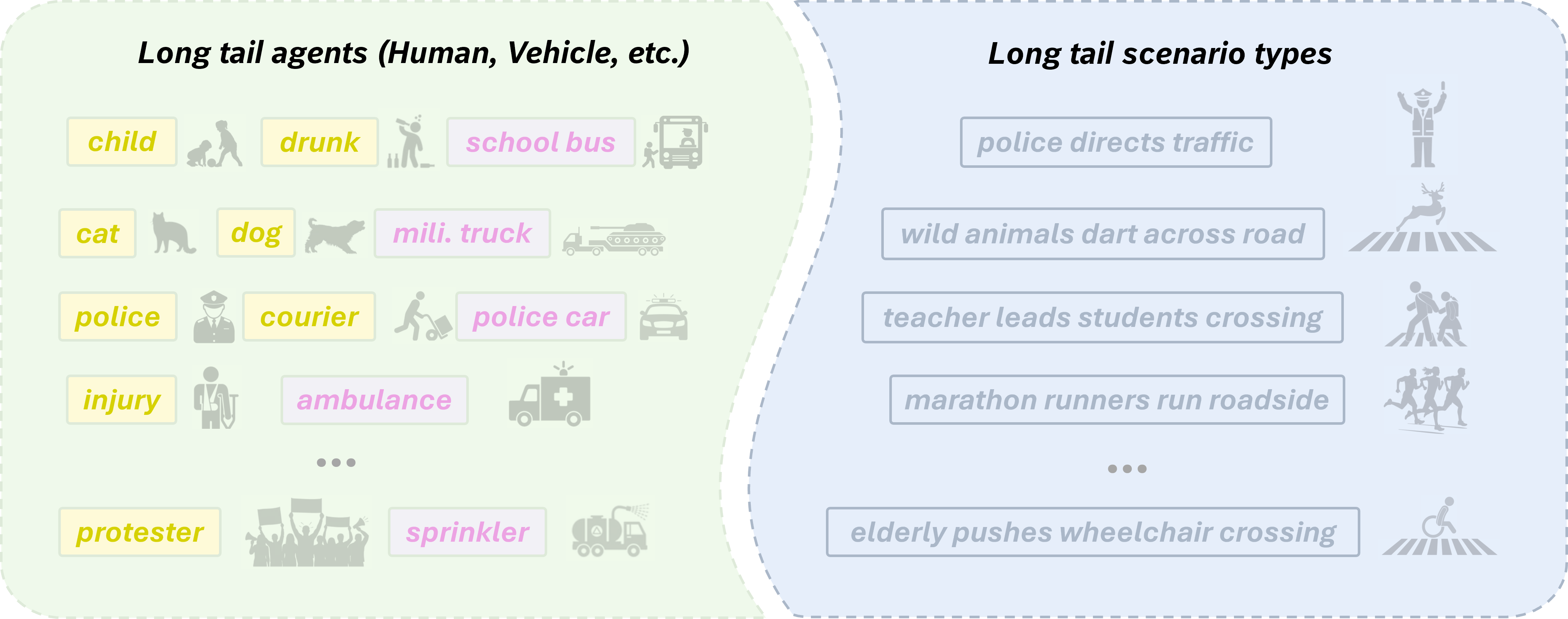}
    \vspace{-0.5em}
    \caption{
    Overview of long-tail agents and scenario types in \dataset.
    }
    \label{fig:teaser}
    \vspace{-0.em}
\end{figure}

To enable such simulation, we construct long-tail and semantically rich scenarios, named the \dataset dataset, on top of real-world driving scenes from the nuPlan dataset. \dataset is designed as a closed-loop planning task where each scenario evaluates whether an ego planner can complete the task safely while interacting with controlled agents. \dataset comprises over 50 scenario types, each involving multiple interactive agents with diverse language instructions. By placing agents at appropriate positions in different traffic scenes, we finally produce over 230 scenarios. \Cref{fig:teaser} provides representative long-tail agents and scenario types in \dataset.

In summary, the main contributions of this paper are:
\begin{itemize}
    \item We propose an agent-driven simulation framework for autonomous driving evaluation, where surrounding road participants are controlled by instruction-following agents rather than predefined rules or log replay, enabling realistic and interactive closed-loop behaviors.

    \item We propose the \dataset dataset, which comprises long-tail and semantically rich scenarios for planning, by adding agents with diverse language instructions to real-world driving scenes.

\end{itemize}

\section{Related Work}

\paragraph{Driving simulators and benchmarks.}
Simulation is a cornerstone for developing and evaluating autonomous driving systems. Classical simulators such as CARLA~\cite{dosovitskiy2017carla} provide controllable synthetic environments, while data-driven simulators aim to improve realism and scale by grounding simulation in real logs, such as Waymax~\cite{gulino2023waymax}, and by providing reusable scenario corpora, such as ScenarioNet~\cite{li2023scenarionet}. Large-scale benchmarks further standardize closed-loop evaluation, with nuPlan~\cite{caesar2021nuplan,karnchanachari2024nuplan} and follow-ups that improve reactivity and long-tail stress testing (nuPlan-R~\cite{peng2025nuplan_R}, interPlan~\cite{hallgarten2024interplan}).
Complementary directions include non-reactive or pseudo-simulation protocols (NAVSIM~\cite{dauner2024navsim}, pseudo-simulation~\cite{cao2025navsimv2}), CARLA-based suites (Bench2Drive~\cite{jia2024bench2drive}, Bench2Drive-R~\cite{you2024bench2drive_R}), and datasets/challenges emphasizing interaction and reasoning (WOSAC~\cite{montali2023waymo_sim_agents}, WOMD-Reasoning~\cite{WOMD_Reasoning}, WOD-E2E~\cite{xu2025waymo_e2e}, V2X-Seq~\cite{yu2023v2x_seq}). Generative and real-to-simulation benchmarks further expand evaluation coverage (DriveArena~\cite{yang2025drivearena}, DriveE2E~\cite{yu2025drivee2e}, DrivingDojo~\cite{wang2024drivingdojo}).

\paragraph{Traffic simulation.}
The advent of transformer architectures and generative models has significantly advanced data-driven traffic simulation, particularly in terms of realism and diversity. Modeling traffic via next-token prediction~\cite{wu2024smart, zhang2025trajtok, peng2025infgen} has proven effective, scalable, and flexible, leveraging the strengths of the transformer architecture. Meanwhile, modeling vectorized traffic through diffusion models~\cite{rowe2025scenario, tan2025scenediffuser} has been shown to enhance the diversity of generated scenes. Beyond fidelity and diversity, controllability has emerged as a crucial objective in traffic simulation. While methods based on adversarial behavior~\cite{ransiek2024adversarial} and reinforcement learning~\cite{rowe2024ctrl} have been explored, recent works have integrated language models into tokenized traffic modeling~\cite{tan2024promptable} to achieve highly flexible control.

\paragraph{Simulation using LLMs.}
Foundation models have recently been explored as controllable interfaces for intent modeling and interaction. HumanSim~\cite{zhou2024humansim} studies language-model-based multi-agent driving behaviors of vehicles, and CitySim~\cite{bougie2025citysim} leverages language-model-based agents for long-horizon urban simulation. Related studies apply language models to structured scenario generation and planning abstractions~\cite{li2025crashagent,xie2025llm_world_model,xietravelplanner}. Our work focuses on autonomous driving planning task by introducing instruction-following interactive agents in nuPlan scenarios.

\paragraph{Planning approaches.}
Rule-based planners remain strong baselines due to interpretability and safety priors, with IDM~\cite{treiber2000idm} and simple rule-based policies~\cite{dauner2023pdm_hybrid_pdm_closed} widely used in benchmarks. Learning-based methods span imitation learning, reinforcement learning, and generative planning, achieving strong closed-loop performance~\cite{scheel2022urban_driver,cheng2024plantf,cheng2024pluto,renz2022plan_cnn,hallgarten2023GC_PGP} and addressing interaction, robustness, and multimodality~\cite{huang2023gameformer,liu2024reasoning_behavioral,chen2023interactive_joint,zhang2023cat,lu2023imitation_not_enough,zheng2025diffusion_planner,hu2024solving_planning}. Language-model-based planners further enable high-level reasoning and controllable planning interfaces~\cite{zheng2024planagent,chen2024asynchronous_llm,sharan2023llm_assist,sima2024drivelm,huangdrivegpt,chen2025drivinggpt,li2023towards_knowledge}.

\section{Method}

\subsection{Overview}

\begin{figure*}[!ht]
    \centering
    \includegraphics[width=0.95\linewidth]{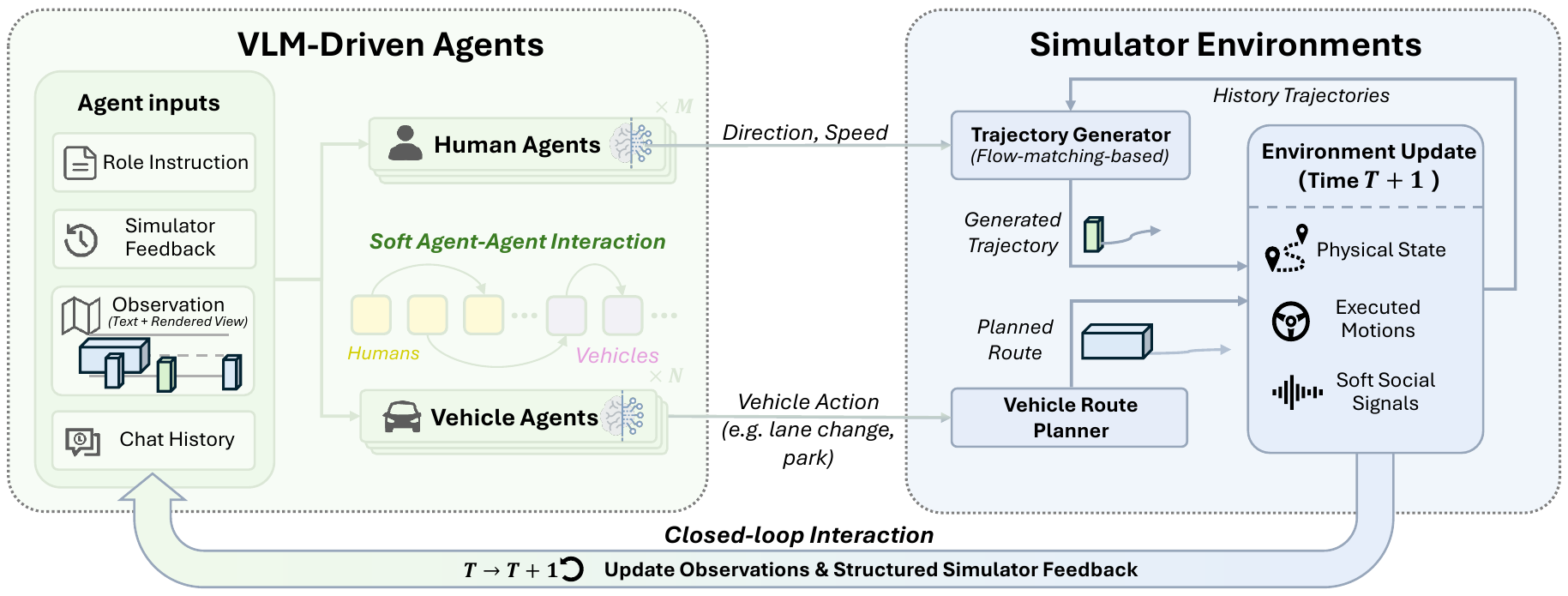}
    \caption{Overview of our agent-driven simulation framework. At each query step, the simulator constructs an agent-centric observation, feedback, and an agent-centric third-person view for each controlled agent. The agents output constrained actions and a structured motion command, which are validated and executed by simulators to produce physically grounded behavior and new feedback events for closed-loop interaction.}
    \label{fig:framework}
\end{figure*}

We consider closed-loop simulation of an autonomous vehicle (ego) interacting with surrounding road participants. Let $s_t$ denote the simulator state at time $t$ and $\mathcal{A}$ the set of non-ego agents. Our goal is to endow a subset of agents $\mathcal{A}_{\text{ctrl}} \subseteq \mathcal{A}$ with intentional, instruction-following, and interactive behaviors, while maintaining physical plausibility.

To build physically plausible simulation, we extend the nuPlan simulator with a set of high-level actions. For human agents, these actions include entering or exiting vehicles, picking up or putting down objects, etc. For vehicle agents, the actions cover behaviors such as parking or starting to drive, and changing lanes. These high-level actions are translated by the simulator into low-level physical constraints and state transitions. \Cref{fig:framework} illustrates the overall framework.

\subsection{Agent Interface: Inputs}
\paragraph{Inputs.}
For each controlled agent we query an agent policy conditioned on four inputs:
\begin{itemize}
    \item \textbf{Role instruction} describing the agent's identity, intent, and constraints.
    \item \textbf{Simulator feedback} reporting the outcomes of the previously executed action, such as collision notifications or failed interactions.
    \item \textbf{Agent local observation} comprising both a structured text summary of the agent-centric scene context at time $t$ and an agent-centric view (nearby agents, IDs, map context, object geometry, and recent history trajectories).
    \item \textbf{Chat history} capturing the agent's dialogue context across time.
\end{itemize}
We write the agent policy as $a_t^i \sim \pi(\text{prompt}_i,\; f_t^i,\; o_t^i,\; m_t^i)$, where $\text{prompt}_i$ is the role instruction, $f_t^i$ is the simulator feedback, $o_t^i = (o_t^{i,\text{text}}, o_t^{i,\text{img}})$ is the agent local observation comprising a structured text channel and a rendered visual channel, and $m_t^i$ is the chat history. In our implementation, human agents are instantiated with a vision-language model so that they can reason over both channels jointly. The simulator provides explicit feedback about the outcome of each executed action, which the agent can leverage to adjust its future decisions over time.

\paragraph{Simulator feedback.}
We represent $f_t^i$ as structured outcome events emitted by deterministic executors. This explicit feedback channel enables controlled agents to perform closed-loop correction, such as retrying after a failed interaction, yielding when blocked, or choosing an alternative maneuver.

\paragraph{Agent local observation.}
We construct $o_t^i$ from two complementary channels.
The \emph{structured text channel} $o_t^{i,\text{text}}$ summarizes the agent-centric scene, including the agent state, nearby agents, map context, and recent events.
The \emph{rendered visual channel} $o_t^{i,\text{img}}$ provides an agent-centric view with semantic colors, visible IDs, map elements, and recent history trajectories.

\subsection{Agent Interface: Outputs}

\paragraph{Human agent outputs.}

\emph{Movement decision.}
The agent outputs a structured motion command consisting of one or two WASD axes, an axis ratio, a walking speed, and optionally a target object or agent. These fields specify a concrete local movement direction and speed, and the simulator converts them into trajectories using a flow-matching trajectory generator.

\emph{Human action.}
Human outputs may also include an optional high-level action.
When an action is accepted, the simulator updates the corresponding constraints to realize state transitions. Picking up attaches the object to the human so that it follows the human's motion; entering a vehicle attaches the human to the vehicle so that the human follows the vehicle.

\emph{Soft agent-agent interaction.}
Human agents can also emit soft signals such as gestures, posture changes, or short utterances. The simulator converts these signals into structured stimulus events for nearby agents, without directly changing their physical states.

\paragraph{Vehicle agent outputs.}
Vehicle agents output high-level maneuvers with a small number of parameters, rather than trajectories or low-level controls. The simulator executes these commands with its native route planner, which improves safety and map compliance compared with direct trajectory or control outputs. Detailed implementation of the route planner is provided in the appendix.

\subsection{Flow-Matching Trajectory Generation}
\label{sec:flow_matching_traj_gen}

Given the requested speed $u$, a concrete local movement direction computed from the agent's WASD axis combination and axis ratio, and a short planning horizon $H$, the simulator generates an executable trajectory $\tau_t^i = \{(x_{\ell}, y_{\ell})\}_{\ell=1}^{L}$, where the trajectory contains $L$ uniformly spaced waypoints over $H$.

We use a conditional flow-matching trajectory generator for this conversion. The simulator first resolves the discrete WASD axes and their ratio into a continuous local direction vector. The generator then receives the agent's recent history trajectory, the requested speed, and this resolved local direction, and samples a short-horizon future trajectory. The generated local trajectory is converted to the world frame and followed by the simulator over the next control horizon. This learned generator produces smooth and realistic short-horizon human motion while preserving a constrained high-level action interface.

\subsection{Simulator Internals}

The simulator constructs compact local observations for each controlled agent from nearby agents, vehicles, objects, map semantics, and structured stimulus events.
For ego planning, we preserve the default nuPlan observation and expose each controlled agent's semantic type together with textual action descriptions.
Vehicle agents are executed through high-level maneuvers with internal route re-planning, and auditory interactions such as honking are represented as explicit simulator events.

\section{Experiments}

\subsection{SemanticPlan Dataset}
\dataset augments real-world nuPlan scenarios with controlled interactive agents, creating semantically rich long-tail scenarios for closed-loop ego planning evaluation.
We construct over 50 scenario types, each containing 1 to 8 manually designed agents with language instructions, and instantiate each type on 3 to 5 different base scenarios.
In total, we produce over 230 scenarios drawn from the nuPlan validation split.
For each scenario type, we first design agent prompts, and then place the agents at appropriate locations within each scenario using a Unity-based scene editor.
Detailed construction procedure and a full list of scenario types are provided in the appendix.

Different scenario types emphasize different aspects of long-tail driving. Some are intrinsically safety-critical and we focus on collisions; others require semantic decisions such as whether to honk, whether to avoid a scenario-specific penalty region, or whether to follow an explicit traffic-police instruction.
We therefore group scenarios into two evaluation tracks.

\subsubsection{Collision-prone Track}
The collision-prone track follows the standard nuPlan closed-loop evaluation setting, where the ego planner receives the usual nuPlan inputs and is scored with safety and progress metrics. We augment only the surrounding-agent behavior, not the planner evaluation protocol.

Collision-prone scenarios primarily test whether the ego can make progress while staying safe and on-road. The score rewards route progress while penalizing drivable-area violations and assigning zero credit to rollouts with ego-at-fault collisions.

\subsubsection{Semantic Track}
The semantic track combines general semantic scenarios and honk-sensitive long-tail scenarios, and we develop LLM-based baselines for this track.

\paragraph{General semantic scenarios.}
The ego should satisfy scenario-specific semantic constraints while completing the task. Most cases use penalty regions and evaluate the ego vehicle's overlap with these regions, while traffic-police cases evaluate explicit stop or lane-change compliance. We combine this penalty with truncated progress. Detailed formulas are provided in the appendix.

\paragraph{Honk-sensitive scenarios.}
Honking can be necessary to make progress, but can also be inappropriate in certain contexts, such as honking at medical personnel or traffic police. We therefore evaluate both progress in honk-required scenarios and a role-weighted penalty for inappropriate honking in honk-penalized scenarios.

\subsection{Experimental Setup}

Our goal is to assess the ability of planners to generalize to unseen and long-tail scenarios, which is critical for safe real-world deployment. Accordingly, all planners are evaluated in a zero-shot setting, where they are trained on the standard large-scale nuPlan training set and are not fine-tuned or adapted on the augmented benchmark scenarios. We summarize full planner comparisons on the collision-prone track in \Cref{tab:collision_track}, and evaluate semantic decision-making with LLM-based planners in \Cref{tab:semantic_track}. To make large-scale planner evaluation practical, the collision-prone track uses pre-generated agent trajectories, while the semantic track keeps real-time agent simulation because semantic decisions such as honking and scenario-specific constraint handling must remain interactive with the evaluated ego planner.

\paragraph{Experimental details.}
We evaluate in closed-loop simulation, where the ego planner interacts with the environment and surrounding agents respond online. Unless otherwise stated, we use a simulation step of $\Delta t=0.1$\,s (10\,Hz). Each candidate trajectory spans a planning horizon of $H=2$\,s with $L=20$ waypoints at the same 0.1\,s resolution. Each controlled agent is queried every $20$ steps (i.e., every $2$\,s) and executes the resulting structured action in-between queries.
The simulation horizon $T$ is set based on scenario type; in particular, for a subset of honk-required scenario types we extend the simulation horizon to allow longer interactive resolution.

We use Qwen3.6-27B with FP8 quantization as the default vision-language agent model for human and vehicle agents. The sampling temperature is set to $0.7$.
We generate $K=3$ stochastic agent rollouts for each scenario instance. For the collision-prone track, these rollouts are saved as trajectory overlays and replayed for every planner, so planner evaluation no longer performs online agent inference.
This preserves stochastic coverage while reducing the expensive generative simulation cost across all baselines. For the semantic track, we keep real-time simulation because the evaluated ego behavior changes the semantic interaction itself, such as whether and when the ego honks or satisfies a scenario-specific constraint.
We compare simulation protocol and report runtime cost in\cref{tab:simulation_time}.
Reported runtime costs are measured on a machine with eight NVIDIA RTX 3090 GPUs and 20 CPU cores.

\begin{table}[h]
\centering
\small
\caption{Simulation protocol and runtime cost measured with the IDM planner over three simulations.}
\label{tab:simulation_time}
\setlength{\tabcolsep}{4pt}
\begin{tabular}{lccc}
\toprule
Benchmark & Human & Vehicle & Eval. time \\
\midrule
nuPlan original & Log replay & IDM & -- \\
Collision-prone track & Pre-generated replay & IDM & 1.5h \\
Semantic track & Real-time agent & Real-time agent & 15.1h \\
\bottomrule
\end{tabular}
\end{table}

\paragraph{Baselines.}

On the collision-prone track, we evaluate both rule-based and learning-based ego planners. For rule-based methods, we include IDM~\cite{treiber2000idm} as a lightweight baseline and PDM Closed as a strong map-aware planning baseline.
For learning-based and hybrid methods, we evaluate UrbanDriver~\cite{scheel2022urban_driver}, GC-PGP~\cite{hallgarten2023GC_PGP}, PlanTF~\cite{cheng2024plantf}, Diffusion Planner~\cite{zheng2025diffusion_planner}, PLUTO~\cite{cheng2024pluto}, and PDM Hybrid~\cite{dauner2023pdm_hybrid_pdm_closed}.

For the semantic track, we focus on planners built on top of IDM. We compare plain IDM, IDM + LLM, and a stop-prioritized IDM + LLM variant. IDM + LLM keeps longitudinal speed IDM-based while querying a Qwen3.6-27B semantic decision channel every $2$\,s for lane changing and scenario-specific interactions such as honking, avoiding penalty regions, or following explicit traffic-police instructions. The stop-prioritized variant enables the same semantic decision channel to issue explicit stop decisions, making the planner more conservative.

\subsection{Baseline Results}

\begin{table}[h]
\centering
\scriptsize
\caption{Performance of popular planners on the collision-prone track.}
\label{tab:collision_track}
\begin{tabular}{l c c >{\columncolor{gray!15}}c}
\toprule
Method & $\text{Prog.}\uparrow$ & $\text{Safe.}\uparrow$ & $\text{Overall}\uparrow$ \\
\midrule
\textit{Rule-based}&  &  &  \\
IDM                & 0.818 & 0.507 & 0.409  \\
PDM Closed         & 0.855 & 0.767 & 0.644  \\
\midrule
\textit{Learning-based \& Hybrid} &  &  &  \\
UrbanDriver        & 0.890 & 0.301 & 0.242  \\
GC-PGP             & 0.604 & 0.432 & 0.258  \\
PlanTF             & 0.861 & 0.459 & 0.380  \\
Diffusion Planner  & 0.935 & 0.425 & 0.373  \\
PDM Hybrid         & 0.856 & 0.774 & \textbf{0.651}  \\
PLUTO              & 0.924 & 0.664 & 0.616  \\
\bottomrule
\end{tabular}
\end{table}

The results on the collision-prone track are shown in \Cref{tab:collision_track}. PDM Hybrid performs best overall, but all planners still show limited safety scores, indicating that these interactive long-tail scenarios remain challenging.

\begin{table}[h]
\centering
\small
\caption{
Zero-shot closed-loop planning performance on the semantic track, which combines general semantic scenarios and honk-sensitive scenarios. All results report the mean over three simulations.}
\label{tab:semantic_track}
\resizebox{\linewidth}{!}{
\begin{tabular}{l c c c c >{\columncolor{gray!15}}c}
\toprule
Method & $\text{Gen. Sem. Prog.}\uparrow$ & $\text{Gen. Sem. Penalty}\downarrow$ & $\text{Honk Prog.}\uparrow$ & $\text{Honk Penalty}\downarrow$ & $\text{Overall}\uparrow$ \\
\midrule
IDM                       & 0.580 & 0.574 & 0.499 & \textbf{0.000} & 0.353 \\
IDM + LLM                 & \textbf{0.600} & 0.561 & \textbf{0.562} & 0.024 & \textbf{0.389} \\
IDM + LLM (stop)          & 0.395 & \textbf{0.146} & 0.335 & 0.004 & 0.309 \\
\bottomrule
\end{tabular}
}
\end{table}

The semantic-track results in \Cref{tab:semantic_track} show that adding the LLM semantic decision channel improves honk progress and the overall score over plain IDM. The stop-prioritized variant reduces region penalty, but it is overly conservative and loses progress, leading to lower overall performance.

\subsection{Simulation Quality}

We evaluate simulation quality on a subset of scenario types whose intended agent behavior can be checked with deterministic rule-based predicates. This evaluation covers 10 scenario types, including object retrieval, vehicle boarding, parcel or cone relocation, driver interaction, and related interactive behaviors. We define \emph{success} using ordered or counted checks over simulator logs, including command events, environment feedback, parent or state transitions, and proximity checks.

\begin{table}[h!]
\centering
\begin{minipage}[t]{0.35\linewidth}
\centering
\small
\captionof{table}{Simulation quality ablations.}
\label{tab:llm}
\setlength{\tabcolsep}{2pt}
\begin{tabular}{clc}
\toprule
Flow Mat. & Model & Success Rate $\uparrow$ \\
\midrule
\cmark & Qwen3.5-4B  & 0.653 \\
\cmark & Qwen3.5-9B  & 0.708 \\
\xmark & Qwen3.6-27B & 0.792 \\
\cmark & Qwen3.6-27B & \textbf{0.889} \\
\bottomrule
\end{tabular}
\end{minipage}
\hfill
\begin{minipage}[t]{0.61\linewidth}
\centering
\small
\captionof{table}{Human trajectory distribution comparison. Acceleration and jerk are computed from 10\,Hz samples.}
\label{tab:human_traj_distribution}
\setlength{\tabcolsep}{2pt}
\begin{tabular}{lr>{\columncolor{gray!15}}r>{\columncolor{gray!15}}rrr}
\toprule
Source & Speed & Accel. & Jerk & Turn & Direction \\
 & m/s & m/s$^2$ & m/s$^3$ & deg & Straight/L/R \\
\midrule
nuPlan logs & 0.97 & 2.96 & 50.10 & 22.9 & 0.42/0.25/0.33 \\
Ours w/o FM & 1.86 & 0.90 & 15.57 & 20.4 & 0.74/0.12/0.14 \\
Ours & 2.04 & 2.90 & 47.34 & 26.1 & 0.54/0.23/0.23 \\
\bottomrule
\end{tabular}
\end{minipage}
\end{table}

Without flow matching, the agent outputs future waypoints directly.
We also evaluate different Qwen3.5 agent models with FP8 quantization to measure sensitivity to model size.
As shown in \Cref{tab:llm}, flow matching yields a higher success rate than direct waypoint generation, and larger agent models have more stable performance.

\Cref{tab:human_traj_distribution} further shows that flow matching produces acceleration, jerk, and direction statistics closer to nuPlan logs, while direct waypoint generation produces overly straight motion.

\subsection{Vehicle Action Comparison}
We compare three vehicle-agent interfaces, including high-level actions executed by the simulator's route planner, direct low-level control, and direct waypoint generation followed by a two-stage controller. As shown in \Cref{tab:vehicle}, high-level actions substantially reduce collisions and improve drivable compliance, because feasibility and kinematic constraints are handled by the simulator rather than the agent model.

\begin{table}[h]
\centering
\begin{minipage}[t]{0.48\linewidth}
\centering
\small
\captionof{table}{High-level vehicle actions vs.\ low-level controls. We report a no-collision score and the drivable-area compliance score of agent controlled vehicles.}
\label{tab:vehicle}
\setlength{\tabcolsep}{2pt}
\begin{tabular}{lcc}
\toprule
Action interface & No Col. $\uparrow$ & Driv. $\uparrow$ \\
\midrule
Low-level traj. & 0.285 & 0.442 \\
Low-level ctrl. & 0.507 & 0.697 \\
High-level actions & \textbf{0.905} & \textbf{0.956} \\
\bottomrule
\end{tabular}
\end{minipage}
\hfill
\begin{minipage}[t]{0.48\linewidth}
\centering
\small
\captionof{table}{Effect and penalty of honking. Here $\text{Prog.}$ measures progress of honk-required scenarios, honk penalty is evaluated on honk-penalized scenarios.}
\label{tab:honk}
\setlength{\tabcolsep}{2pt}
\begin{tabular}{lccc}
\toprule
Strategy & Prog. $\uparrow$ & Pen. $\downarrow$ & \scorehonk$\uparrow$ \\
\midrule
Never honk & 0.457 & \textbf{0.000} & 0.452 \\
Always honk & \textbf{0.913} & 0.201 & \textbf{0.493} \\
\bottomrule
\end{tabular}
\end{minipage}
\end{table}

\begin{figure*}[!t]
  \centering
  \begin{minipage}[t]{0.48\textwidth}
    \vspace{0pt}          
    \centering
    \begin{subfigure}[t]{\linewidth}
      \centering
      \includegraphics[width=\linewidth,trim=1mm 1mm 1mm 1mm,clip]{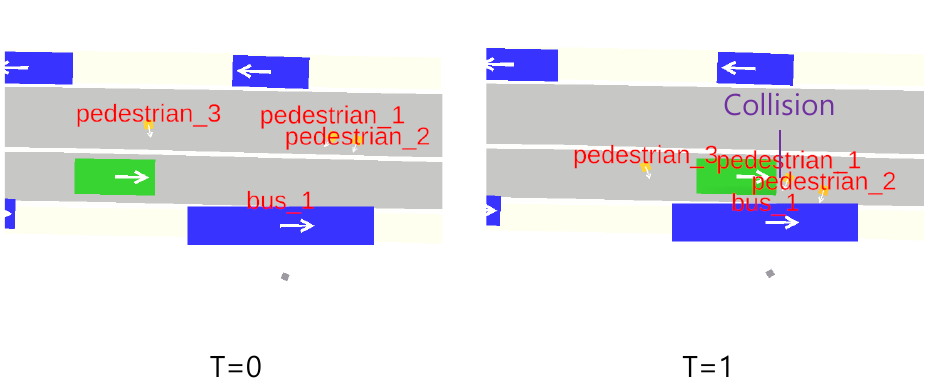}
      \caption{Pedestrians suddenly dash across the roadway to catch an approaching bus.}
    \end{subfigure}
    \vspace{0.08cm}
    \begin{subfigure}[t]{\linewidth}
      \centering
      \includegraphics[width=\linewidth,trim=1mm 1mm 1mm 1mm,clip]{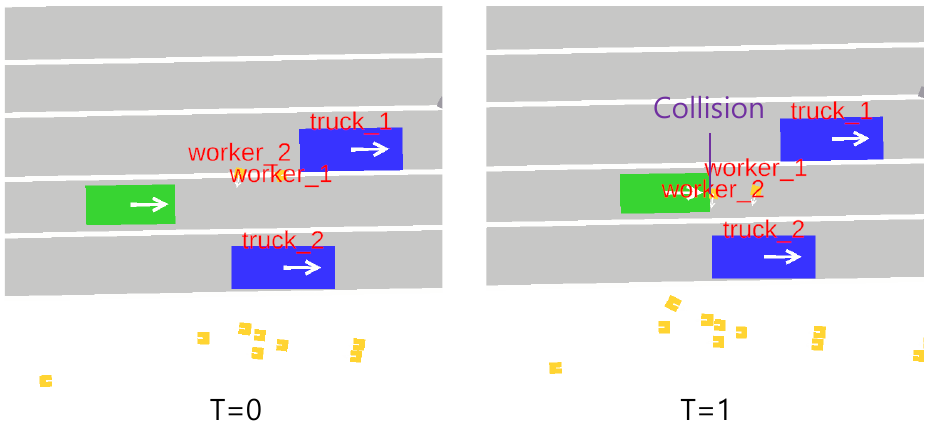}
      \caption{Scattered bicycles partially block the sidewalk and force pedestrians toward the driving lane.}
    \end{subfigure}
    \captionof{figure}{Representative scenarios in the collision-prone track.}
    \label{fig:collision}
  \end{minipage}%
  \hfill
  \begin{minipage}[t]{0.48\textwidth}
    \vspace{0pt}          
    \centering
    \includegraphics[width=\linewidth,trim=3mm 3mm 3mm 3mm,clip]{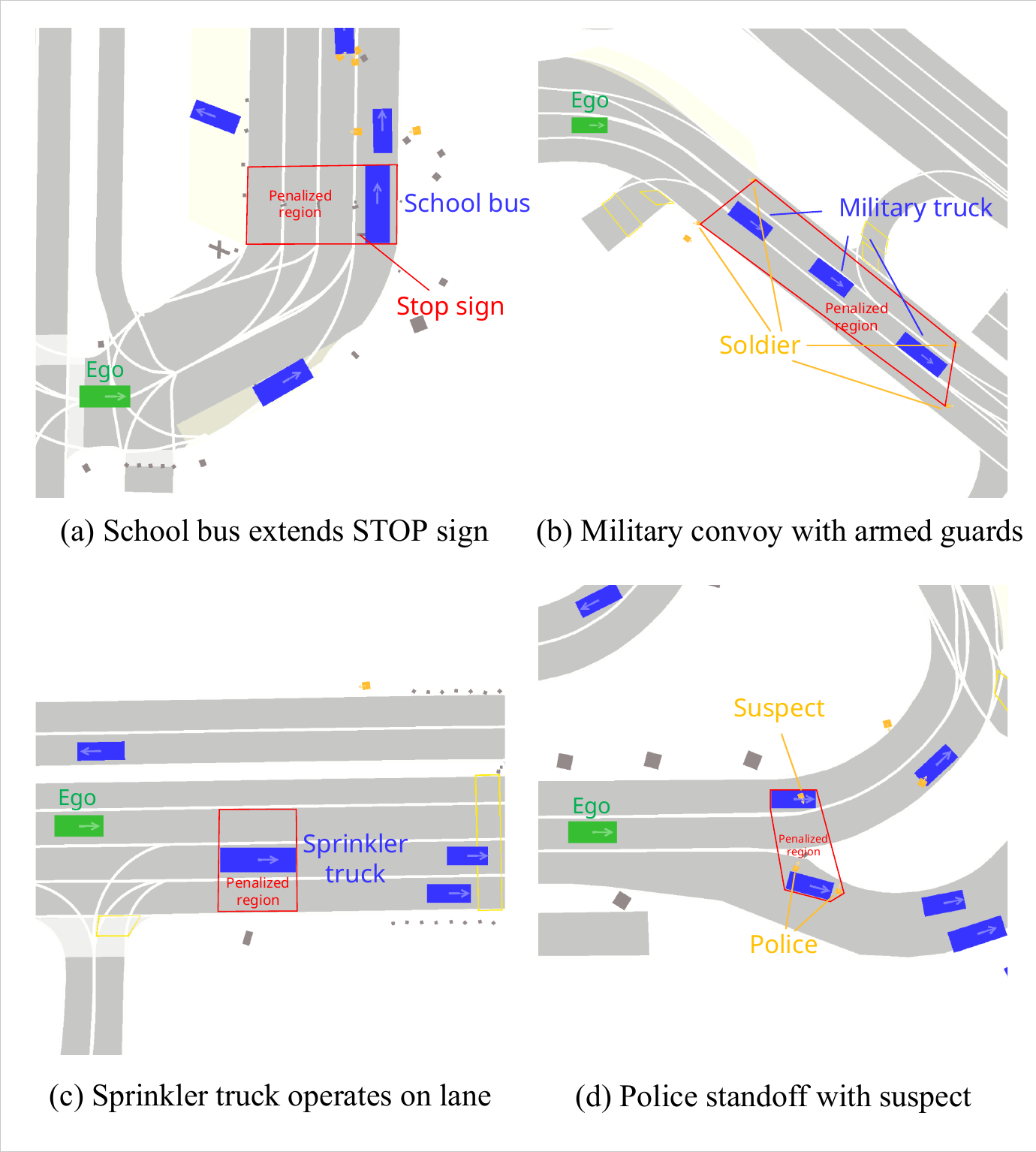}
    \captionof{figure}{Representative scenarios in the semantic track.}
    \label{fig:region}
  \end{minipage}
\end{figure*}

\subsection{Honking Analysis}
We evaluate two extreme ego honking strategies with IDM control: always honk and never honk. As shown in \Cref{tab:honk}, always honking improves progress in honk-required scenarios, but introduces a nonzero penalty in honk-penalized scenarios; the resulting \scorehonk captures this progress--appropriateness trade-off.

\subsection{Qualitative Examples}

\paragraph{Collision-prone scenarios.}
We present qualitative visualizations of collision-prone scenarios using IDM as the planner to illustrate the safety challenges faced by ego planners in closed-loop evaluation, as shown in \Cref{fig:collision}.

\paragraph{Semantic scenarios.}
Semantic scenarios evaluate whether an ego planner can make progress while satisfying a scenario-specific semantic constraint. Most constraints are spatial penalty regions, while traffic-police cases use explicit stop or lane-change compliance. We present four representative scenario types, as shown in Figure~\ref{fig:region}.
In these cases, the desired behavior is to detour around the region when feasible, yield and wait until the region clears, or follow the explicit instruction. For scenarios where stopping is the intended behavior, we adjust the ego goal specification used to compute progress so that waiting at a safe location is considered successful task completion.

\section{Conclusion}

In summary, we presented an agent-driven closed-loop simulation framework for autonomous driving, where selected surrounding participants are controlled by instruction-following agents through a constrained, structured action interface.
Building on this framework, we introduce \dataset, a benchmark of semantically rich long-tail planning scenarios constructed by augmenting real nuPlan scenes with multiple interactive agents following diverse role instructions,
highlighting the gap between standard benchmark performance and robust behavior in rare, long-tail scenarios.
We aim to provide a useful platform for studying closed-loop planning in challenging interactive and long-tail driving scenarios.

\clearpage


\bibliography{example_paper}  

\clearpage
\appendix

\section{Simulation Details}
\label{app:simulator_details}

\paragraph{Ego observation.}
For ego planning, we preserve the default nuPlan observation, including map, route, and tracked-object states.
For the semantic track, we additionally expose each controlled agent's identifier and brief semantic actions.
Each nearby human agent emits a short textual description of its currently visible action, such as ``waving to stop a car'', produced as part of its structured output at every query and included in the ego observation at the next step.

\paragraph{Semantic-track evaluation.}
The semantic track keeps the controlled agents in the loop during planner evaluation, so agent inference runs online together with the ego planner. We deploy the Qwen3.6-27B model in FP8 precision with vLLM~\cite{vLLM} on a machine with eight NVIDIA RTX 3090 GPUs. For learning-based ego planners, we limit the vLLM KV cache memory budget so that one GPU remains available for hosting the planner model. We enable vLLM prefix caching to reduce repeated computation across multi-turn agent conversations. The maximum number of concurrently running agents is set to 16. Under this setup, one full semantic-track evaluation with the IDM planner takes 15.1 hours.

\paragraph{Route planner.}
Vehicle agents are controlled through high-level maneuvers rather than low-level actuation. The agent outputs abstract vehicle actions, such as lane changes and parking, which are realized by the route planner through re-planning and native vehicle dynamics. This design avoids direct trajectory or control generation by the agent and significantly improves robustness in dense interactive scenes.

The route planner is implemented on top of the nuPlan lane graph and IDM vehicle model. Each controlled vehicle maintains an IDM route as a sequence of lane or lane-connector segments and a centerline path sampled from their baseline paths. During normal driving, the planner extends the route when the remaining path is short by selecting valid outgoing map edges under the current traffic-light state, preferring the lowest-curvature outgoing edge to continue straight when multiple options are available. For commands that change the route, such as lane change, lateral offset, start driving, or drive-to-lane, the simulator first identifies a target lane segment from the map API. It then projects the current vehicle pose onto the target segment, computes an insertion point after a command-dependent forward distance, and builds a short S-curve approach from the current pose to that insertion point. The final path concatenates this approach with the target lane baseline after the insertion point, and IDM longitudinal control propagates the vehicle along the planned path while accounting for nearby occupancy. If no feasible current lane, adjacent lane, target lane, or sufficient road space is found, the command is rejected and the failure reason is returned as simulator feedback to the agent.

When a human agent is designated as a driver, we bind the driver's identity to the corresponding vehicle over time. The driver can then issue vehicle-level actions through the same structured action interface, enabling compositional behaviors such as driving, parking, exiting the vehicle, and interacting with the environment within a single coherent agent lifecycle.

\paragraph{Simulator feedback.}
Simulator feedback is represented as structured outcome events emitted by deterministic executors. For human agents, feedback includes whether interaction attempts succeed or fail, such as entering or exiting a vehicle and picking up, putting down, or pushing an object, together with reasons such as invalid targets, insufficient proximity, or infeasible preconditions. It also includes contact warnings when a proposed motion touches a blocking entity. For vehicle agents, feedback includes maneuver status signals such as started, completed, or failed, along with failure causes such as no feasible lane or route, being blocked by nearby entities, or timing out.

\paragraph{Honking channel.}
To model honking signals commonly used in real-world driving, we introduce an explicit honking event triggered by the ego vehicle. When activated, the simulator injects an honking stimulus and delivers it to nearby agents. The honking event is treated as a structured stimulus in the agents' observations, enabling instruction-conditioned and closed-loop reactions to auditory signals.

\subsection{Flow-matching-based Trajectory Generation}
\label{app:traj_gen}

\paragraph{Model.}
We generate each human agent's future trajectory with a conditional flow-matching model adapted from \cite{tan2026flow}, where we retain the flow-matching trajectory backbone but modify the conditioning inputs for our setting. Given the agent's recent motion history ($2.0$\,s), a target direction, and a target speed, the model directly produces a future trajectory over a $2.0$\,s horizon ($20$ steps at $10$\,Hz) by integrating the flow ODE from a Gaussian sample. The target direction and speed are obtained from the agent's movement intent (\Cref{app:human_output_format}).

\paragraph{Training data.}
We extract human pedestrian trajectories from the nuPlan validation split (subsampled to $10$\,Hz in the agent-local frame), since the scenarios in \dataset are constructed on top of nuPlan validation logs. After balancing across speed and direction bins, we sample $\sim$$1.4$M human trajectories for training and a disjoint set of one-tenth as many trajectories (from non-overlapping logs) for testing.

\paragraph{Evaluation.}
We evaluate on the held-out test set described above. We assess two properties. \emph{Reconstruction}: conditioning on the ground-truth direction and speed, the generated trajectory attains an average displacement error (ADE) of $0.151$\,m and a final displacement error (FDE) of $0.203$\,m against the true future. \emph{Controllability}: issuing a target direction offset from the agent's motion history by $\{0,\pm45,\pm90,\pm135,180\}^\circ$, we measure the Direction Error between the generated trajectory's final direction and the commanded direction, together with the fraction of samples within $15^\circ$ (hit@$15^\circ$). As summarized in \Cref{tab:traj_gen_eval}, the generator follows the commanded direction closely (mean Direction Error $3.68^\circ$, hit@$15^\circ \geq 97\%$), and the largest errors occur on sharp turns ($\pm90^\circ/\pm135^\circ$) that most oppose the motion history.

\begin{table*}[h]
\centering
\small
\caption{Direction controllability and displacement error of the human trajectory generator on the held-out test set.}
\label{tab:traj_gen_eval}
\setlength{\tabcolsep}{6pt}
\begin{tabular}{lcccccc}
\toprule
Metric & $0^\circ$ & $45^\circ$ & $90^\circ$ & $135^\circ$ & $180^\circ$ & Mean \\
\midrule
Direction Error (deg) $\downarrow$ & 1.87 & 2.66 & 4.64 & 4.73 & 3.51 & 3.68 \\
hit@$15^\circ$ (\%) $\uparrow$ & 99.2 & 99.3 & 94.7 & 95.8 & 97.6 & 97.0 \\
\midrule
Average Displacement Error (ADE, m) $\downarrow$ & -- & -- & -- & -- & -- & $0.151$ \\
Final Displacement Error (FDE, m) $\downarrow$ & -- & -- & -- & -- & -- & $0.203$ \\
\bottomrule
\end{tabular}
\end{table*}

\section{Dataset Details}

\subsection{Dataset Construction}
\label{app:dataset_construction}

\begin{figure}[h]
    \centering
    \IfFileExists{fig/dataset_pipeline.pdf}{
        \includegraphics[width=0.95\linewidth]{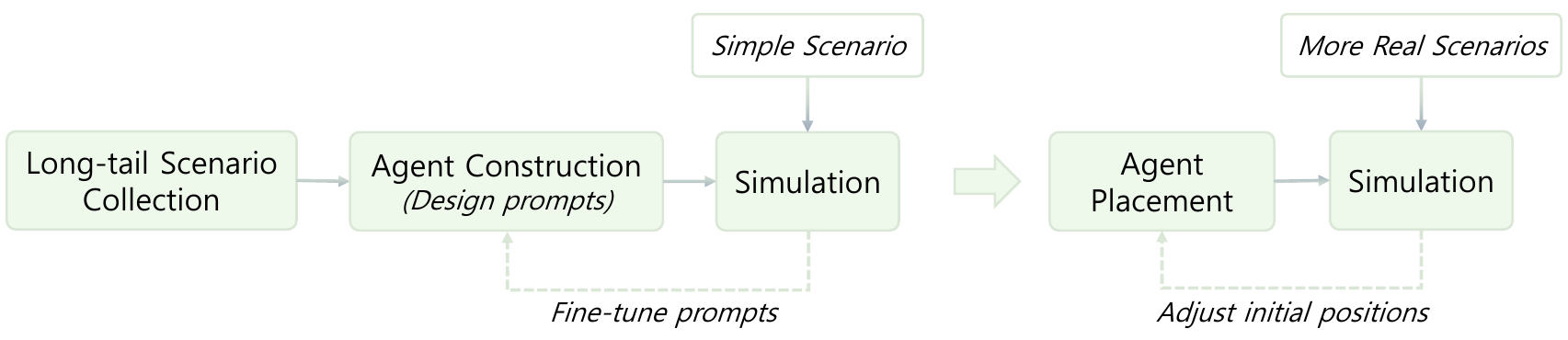}
    }{
        \IfFileExists{fig/dataset_pipeline.png}{
            \includegraphics[width=0.95\linewidth]{fig/dataset_pipeline.png}
        }{
            \fbox{\parbox{0.9\linewidth}{\centering Placeholder for dataset construction pipeline.}}
        }
    }
    \caption{Dataset construction pipeline.}
    \label{fig:dataset_pipeline}
\end{figure}

Here we show the detailed procedure of our dataset construction, as shown in \Cref{fig:dataset_pipeline}. Annotators first collect approximately 80 possible long-tail driving scenario types based on their driving experience. We then filter out types that are difficult to instantiate in our simulator, especially visual long-tail cases that primarily depend on appearance rather than agent behavior, such as snow-covered roads, potholes, or small road debris. This filtering leaves more than 50 interaction-focused long-tail scenario types.

For each remaining scenario type, annotators specify the participating agents (1 to 5 agents for each scenario type) and design the corresponding agent prompts. They then use a Unity3D scene editor to place the agents on a simple scenario and run closed-loop simulations to inspect whether the resulting behavior matches the intended long-tail semantics. The prompts  are iteratively refined until the simulated behavior is stable and semantically aligned with the target scenario type. After this process, the prompt for each agent in a scenario type is fixed.

Finally, annotators instantiate each scenario type on 3 to 8 real scenarios from the nuPlan validation set, choosing plausible agent positions that match the map layout and surrounding traffic context. They inspect the resulting simulations and refine the initial placements when necessary. This process produces more than 50 scenario types and over 230 scenarios in total.

\subsection{Scenario Type List}

Full list of scenario types is shown in \Cref{tab:scenario_types}. The \emph{collision-prone} category forms the collision-prone track. The semantic track consists of \emph{general semantic} and \emph{honk-sensitive} scenarios, where honk-sensitive scenarios are further divided into honk-required and honk-penalized cases.

\label{app:scenario_types}
\begin{table*}[h]
\centering
\small
\caption{Full list of scenario types in \dataset. Some scenario types belong to both the collision-prone track and the semantic track.}
\label{tab:scenario_types}
\setlength{\tabcolsep}{10pt}

\resizebox{\linewidth}{!}{
\begin{tabular}{ll}
\toprule
\multicolumn{2}{l}{\textbf{Collision-prone track}} \\
\midrule
Elderly pushes wheelchair across road & Children on scooters repeatedly cross the road \\
Pedestrian uses phone and drifts & Pedestrian chases windblown hat into road \\
Child breaks free from parent and runs into road & Walked dog chases squirrel into lane \\
Children play on road & Child runs to retrieve ball \\
Drunk person staggers into road & Courier parks tricycle then crosses with package \\
Delivery rider weaves fast through traffic & Salesperson solicits cars in roadway \\
Police question then chase suspect & Wedding convoy moves slowly with followers \\
Marathon runners race along roadside at varied speed & Crowd spreads into road as event ends \\
Crowd gathers at bike share station to scan and ride off & Pedestrian carries long pole sweeping adjacent lane \\
People dry crops on half roadway & Photographer shoots on road with cables \\
Courier places parcels onto roadside delivery motorcycle & One pedestrian crosses on red and others follow \\
Pedestrian dashes across to catch bus & Rider drags dead e-scooter into lane \\
Pedestrians weave between multiple buses at stop & Workers move cones to clear driving space \\
Drivers talk in lane after minor crash & Students cross and some stop to tie shoe \\
Pedestrians push strollers and carts while crossing abreast & Pedestrians file out of construction gap \\
Overwide foam-box cargo on tricycle spills into adjacent lane & Aid station runners converge and spread into lane \\
Food carts with large canopies cluster at intersection & Workers move between two parked trucks \\
Scattered bikes force pedestrians to detour into lane & Deer dart across road erratically \\
\midrule
\multicolumn{2}{l}{\textbf{Semantic track}} \\
\midrule
\multicolumn{2}{l}{\quad\textit{General semantic}} \\
\midrule
Traffic police direct vehicles to stop & Traffic police direct vehicles to change lanes \\
Children dart out around stopped school bus & Military convoy with armed soldiers \\
Workers move between two parked trucks & Sprinkler truck operates slowly in center lane \\
Long large truck drives ahead & Police standoff with suspect \\
Vehicle on fire and bystanders panic & Teacher leads students crossing \\
Ambulance and medics treat injured & \\
\midrule
\multicolumn{2}{l}{\quad\textit{Honk-required}} \\
\midrule
Street performer draws crowd & Bystanders block road at crash scene \\
Spectators line up and wait & Residents do square dancing \\
Promotional event draws lingering crowd & Tourists swarm around tour bus to take photos \\
Cat sleeps on the road & \\
\midrule
\multicolumn{2}{l}{\quad\textit{Honk-penalized}} \\
\midrule
Municipal workers repair manhole and cables & Scavenger pushes overloaded cart slowly \\
Traffic police handle accident and direct flow & Protesters march with banners \\
Religious pilgrims process and kneel & Vendor cart overturns and pedestrians help pick up goods \\
Injured rider lies by fallen e-bike as bystanders gather & Animal rights protesters block livestock truck \\
\bottomrule
\end{tabular}
}

\end{table*}

\subsection{Metric Details}
\label{app:metric_details}

\paragraph{Collision-prone score.}
Let $c_{\text{coll}}\in\{0,1\}$ indicate whether an ego-at-fault collision occurs, and let $o\in[0,1]$ denote drivable-area compliance. We define the combined safety-and-compliance score as $s = (1-c_{\text{coll}})\cdot o$, and compute the final collision-prone score as $\scorecollision = \mathbb{E}\big[p\cdot s\big]$, where $p$ is route progress.

\paragraph{General semantic score.}
For general semantic scenarios, we truncate at the first safety violation time $t_{\text{viol}}$ and compute progress $\tilde{p}$ at $\min(t_{\text{viol}}, T)$. The score uses scenario-specific penalties. For region-based cases, the per-rollout penalty aggregates ego-footprint overlap with the scenario-specific penalty region:
\begin{equation}
r = \mathrm{clip}_{[0,1]}\Big(\frac{1}{\tau}\sum_{t=0}^{\min(t_{\text{viol}}, T)}\frac{\mathrm{area}(\mathcal{R}_t \cap E_t)}{\mathrm{area}(E_t)}\Big),
\end{equation}
where $\mathcal{R}_t$ is the penalty region at time $t$, $E_t$ is the ego footprint, and $\tau$ is a scenario-specific tolerance coefficient. Larger $\tau$ allows brief overlap, while smaller $\tau$ strongly penalizes any overlap.
For traffic-police cases, the command penalty $r$ is zero only if the ego follows the corresponding stop or lane-change instruction.
The final score is $\scoreregion = \mathbb{E}\big[\tilde{p}\cdot (1-r)\big]$.

\paragraph{Honk-sensitive score.}
For each honk-penalized agent $j$, we count the number of honking events directed at that agent, denoted $n_j$. Each such agent has a role-dependent penalty weight $w_j\in[0,1]$ that reflects social appropriateness: protected roles such as police officers or injured riders receive high penalty weight, maintenance workers receive medium weight, and ordinary pedestrians receive no penalty. With decay factor $\gamma=0.7$, we compute the honking penalty as $h=\mathrm{clip}_{[0,1]}\big(\sum_j w_j \cdot (1-\gamma^{n_j})\big)$. The final honk-sensitive score is $\scorehonk = \mathbb{E}\big[\tilde{p}\cdot (1-h)\big]$.

\subsection{Agent Output Format}
\label{app:agent_output_format}

This section documents the structured YAML output formats of the agents.

\paragraph{Human agent output format.}\mbox{}\par
\label{app:human_output_format}

\begin{lstlisting}[
  language=YAML,
  basicstyle=\ttfamily\scriptsize,
  frame=single,
  breaklines=true,
  columns=fullflexible
]
# Brief analysis.
analysis: ""

# Rough movement intent for the next about 2 seconds.
move:
  # If you want to move directly toward a visible target agent, write its id here; otherwise null.
  target_id:
  # Optional stop distance when approaching target_id; otherwise null.
  stop_at_distance:
  # Main movement axis. Must be exactly one of W/A/S/D.
  wasd_primary_axis: "W|A|S|D"
  # Optional auxiliary axis for diagonal movement. Use null for pure primary-axis movement.
  # It must be perpendicular to the primary axis, e.g. primary W can use auxiliary A or D.
  wasd_secondary_axis: null  # null | "W" | "A" | "S" | "D"
  # Ratio primary:secondary. Output your own positive numeric ratio, e.g. "1:0", "1:1", "3:1", "5:3".
  wasd_axis_ratio: ""
  # Rough walking speed in m/s. Use 0.0 if you decide not to move now.
  # range from 0.0 to 4.0, do not exceed 4.0
  speed: 0.0

# Predefined simulator action. Set to null if no predefined action is needed.
predefined_action:
  type: "null|enter_vehicle|exit_vehicle|pick_up_object|put_down_object|push_object"
  # For enter_vehicle or exit_vehicle:
  vehicle_id: "<some_vehicle_id>"
  # For pick_up_object|put_down_object|push_object:
  object_id: "<some_object_id>"
  # For put_down_object, put_down_target is required: use "ground" to drop on ground, or an agent id.
  # If it is null/empty, put_down_object will not take effect.
  put_down_target: "<some_agent_id>|ground"

# Soft action outside predefined_action that may affect other agents, such as humans or vehicles.
# action_description: describe the action using a phrase
# action_description will be delivered to the target agents.
# Set soft_action to null if there is no action that affects other agents.
soft_action:
  action_description: ""
  target_agent_ids: []

pose: "stand|walk|run|sprint|sit|kneel|lie|crouch"
# Whether you are stationary at the current moment (true/false).
is_stationary:
# Briefly describe your current visible action in a short phrase.
brief_visable_action: ""

# Optional camera/view control for Stage 2 and future observations.
# Use none by default.
# left_90 means rotate the view counter-clockwise by 90 degrees to look left.
# right_90 means rotate the view clockwise by 90 degrees to look right.
# back_180 means rotate the view by 180 degrees to look backward.
view_rotation: "none|left_90|right_90|back_180"

\end{lstlisting}

\paragraph{Vehicle agent output format.}\mbox{}\par
\label{app:vehicle_output_format}

\begin{lstlisting}[
  language=YAML,
  basicstyle=\ttfamily\scriptsize,
  frame=single,
  breaklines=true,
  columns=fullflexible
]
# Brief analysis.
analysis: ""

# High-level vehicle command executed by the simulator. If the agent does not want to execute any command or wants to keep the current state, set `command: null`.
command:
  # Allowed `type` values:
  # park: Park at `forward_distance` meters ahead; optional `lateral_distance` (meters, left negative / right positive).
  # lane_change: Change lane. Set `direction=left|right`, `lane_change_time` (seconds, 2~10), and `forward_distance` (meters) to indicate where the lane change should finish.
  # start_driving: Start from a parked state and merge into the nearest lane; `forward_distance` indicates the insertion distance along the lane.
  # lateral_offset: Slightly offset within the current lane; set `direction`, `lateral_offset_time` (seconds), and `forward_distance` (meters) to indicate where the offset should end. For an actual lane change, use `lane_change` instead.
  # drive_to_lane: Drive to a specified lane (useful for turns / U-turns).
  # accelerate: Set `target_velocity` (m/s) with `max_accel` (m/s^2, 0.5~3.0).
  # decelerate: Set `target_velocity` (m/s) with `max_decel` (m/s^2, 0.5~3.0).
  # reverse: Reverse for `reverse_distance` meters (only takes effect when current speed is 0). Optional `use_last_path=true` to reverse along the recently driven path.
  # exit_vehicle: Let the human driver exit the vehicle (only when a human driver is inside).
  # honk: Honk once; nearby agents will receive a honk stimulus.
  type: "park|lane_change|start_driving|lateral_offset|drive_to_lane|accelerate|decelerate|reverse|exit_vehicle|honk|null"

  # For park / start_driving / lane_change / lateral_offset:
  forward_distance:

  # For lane_change: smaller => more urgent (higher curvature), larger => more comfortable.
  lane_change_time:

  # For park (optional):
  lateral_distance: null

  # For lane_change / lateral_offset:
  direction: "left|right"

  # For lateral_offset:
  lateral_offset_time:
  offset: 0.5

  # For drive_to_lane:
  lane_id:

  # For accelerate / decelerate:
  target_velocity:
  max_accel:
  max_decel:

  # For reverse:
  reverse_distance: 3.0
  use_last_path: false

  # Whether honking, together with other commands (does not affect the main command).
  honk: false

\end{lstlisting}

\end{document}